\pdfoutput=1

\documentclass[11pt]{article}

\usepackage[]{acl}
\usepackage{listings}
\usepackage{times}
\usepackage{latexsym}
\usepackage{multirow}
\usepackage{float}
\usepackage{graphicx}

\usepackage{mathtools}
\usepackage{algorithm,algpseudocode}

\newcommand{\ie}{\textit{i.e.}}

\usepackage{amssymb}
\usepackage{amsmath}

\usepackage[T1]{fontenc}

\usepackage[utf8]{inputenc}

\usepackage{microtype}

\title{Embedding Hallucination for Few-Shot Language Fine-tuning}

\author{Yiren Jian\thanks{~~Both authors contributed equally to this research.} \\ Dartmouth College \\
        \texttt{yiren.jian.gr@dartmouth.edu} \\
        \AND
        Chongyang Gao$~^*$ \\ Northwestern University \\ \texttt{cygao@u.northwestern.edu} \\ \And
        Soroush Vosoughi \\ Dartmouth College \\ \texttt{soroush@dartmouth.edu} \\  }

\begin{document}
\maketitle
\begin{abstract}
Few-shot language learners adapt knowledge from a pre-trained model to recognize novel classes from a few-labeled sentences. In such settings, fine-tuning a pre-trained language model can cause severe over-fitting. In this paper, we propose an Embedding Hallucination (EmbedHalluc) method, which generates auxiliary embedding-label pairs to expand the fine-tuning dataset. The hallucinator is trained by playing an adversarial game with the discriminator, such that the hallucinated embedding is indiscriminative to the real ones in the fine-tuning dataset. By training with the extended dataset, the language learner effectively learns from the diverse hallucinated embeddings to overcome the over-fitting issue. Experiments demonstrate that our proposed method is effective in a wide range of language tasks, outperforming current fine-tuning methods. Further, we show that EmbedHalluc outperforms other methods that address this over-fitting problem, such as common data augmentation, semi-supervised pseudo-labeling, and regularization. The code will be made available at: https://github.com/yiren-jian/EmbedHalluc. 
\end{abstract}

\section{Introduction}
Fine-tuning a pre-trained language model (LM) on a downstream task with the labeled data has been the de facto approach in many NLP tasks \cite{wang2018glue, BERT}. Conventional fine-tuning has been shown to be effective when a few thousands of labeled examples are available. Data augmentation \cite{wei-zou-2019-eda}, regularization \cite{lee2019mixout} and re-initialization \cite{zhang2021revisiting} further improve the results.

However, the performance drops drastically when the number of examples falls to only a few dozens. Experiments from recent work \cite{gao2021making} have shown that fine-tuning performs poorly in the setting where only 16 examples per class are given. Indeed, tuning a language model with hundreds of millions of parameters (e.g., BERT-large has 300M parameters) with only a few examples inevitably faces the over-fitting problem.

Prior work have proposed regularization methods to overcome this problem \cite{lee2019mixout, zhang2021revisiting}. However, we show in our experiments that these methods fail in extreme data scarce setting. We speculate that the key to solve this issue is by data augmentation.

Current common text data augmentation methods, such as EDA \cite{wei-zou-2019-eda} (which have been used in recent few-shot learning papers \cite{wei-etal-2021-shot, DBLP:journals/corr/abs-2109-08754}) and AEDA \cite{DBLP:conf/emnlp/KarimiR021} operate at the lexical level, which while resulting in human readable texts, lead to limited diversity due to the discrete nature of the lexical space. In this work, we propose to use a generative augmentation method at the \emph{embedding space} for few-shot learning. The underlying hypothesis is that the intra-class relation of the observed examples can be modeled and that this can be learned from a few-samples to hallucinate diverse unseen examples. To be specific, we adapt a conditional Wasserstein Generative Adversarial Network (cWGAN) \cite{pmlr-v70-arjovsky17a} as our hallucinator to hallucinate embeddings of sentences. By observing the real embeddings of examples from the fine-tuning dataset, the cWGAN plays an adversarial game to hallucinate embeddings that can fool the discriminator, while the discriminator is trying to classify the fake embeddings from the real ones. Once the halluciantor is trained, we condition it on labels to generate diverse embeddings at each fine-tuning step. This effectively extends the fine-tuning dataset with diverse embedding-label pairs which carry intra-class variation that can be a useful learning signal for the language learner. 

We evaluate our method, called Embedding Hallucination (Embedhalluc), on 15 tasks and show that it generally improves over recent fine-tuning methods. We further experimentally show the overall superiority of EmbedHalluc when comparing to regularization methods proposed to address the problem of over-fitting during fine-tuning of LMs, such as Mixout \cite{lee2019mixout} and Re-Init \cite{zhang2021revisiting}. Finally, since our method is a form of data augmentation, we also compare EmbedHalluc to a common data augmentation technique EDA, and semi-supervised learning where unlabeled data is already available.

\section{Related Work}
\textbf{Fine-tuning of Language Models}.
Better fine-tuning of language models can be achieved by proper initialization \cite{Dodge2020FineTuningPL}, regularization \cite{lee2019mixout} or prompts \cite{Schick2021ExploitingCF}. Other tricks include bias correction in optimizer and re-initialization of top layers in Transformer \cite{zhang2021revisiting}. Instead of fine-tuning all parameters in a model, other work explore only learning a few vectors \cite{lester-etal-2021-power, li-liang-2021-prefix, guo-etal-2021-parameter} or a few additional parameters \cite{houlsby2019parameter}. 

\noindent \textbf{Hallucination Methods}.
Feature Hallucination of examples is first introduced for visual recognition \cite{hariharan2017low} by meta-learning \cite{wang2018low}, variational inference \cite{luo2021few, lazarou2022tensor}, and adversarial learning \cite{li2020adversarial, ASH}. Label Hallucination \cite{Jian2022LabelHalluc} assigns soft pseudo-labels for unlabelled images to extend the fine-tuning few-shot dataset. 

\noindent \textbf{Learning from limited labeled data} (few-shot learning) in Computer Vision is usually achieved by meta-learning \cite{ren2018meta, ren18l2rw, Jian2020TMT, Jian2021MetaPix} or transfer learning \cite{tian2020rethinking}.
In NLP, few-shot learning has been successfully applied to machine translation \cite{arthaud-etal-2021-shot}, abstract summarizing \cite{fabbri-etal-2021-improving}, question and answering \cite{hua2020few, ram2021few}, and entity recognition \cite{de2021meta, tong-etal-2021-learning, ding2021few}, by meta learning \cite{li2021semi, bansal2020self, sharaf2020meta}, data augmentation \cite{wei-etal-2021-shot, wei-zou-2019-eda, DBLP:conf/emnlp/KarimiR021, jian2022LMSupCon}, and prompts \cite{gao2021making, tam2021improving}.

Our method is a generative data augmentation method in the embedding space. Different from \cite{wei-etal-2021-shot} which uses EDA \cite{wei-zou-2019-eda} to augment examples at the discrete input space, we hallucinate auxiliary examples at the embedding space. Our method shares similarity to FDA \cite{kumar-etal-2019-closer}, which is also a generative data augmentation method, but at the feature space. Also, different from FDA which is focused on two intent classification tasks, our method can be applied to a wide-range of NLP task as shown by our experiments on 15 diverse tasks.

\begin{figure}[!t]
\centering
\includegraphics[width=0.35\textwidth]{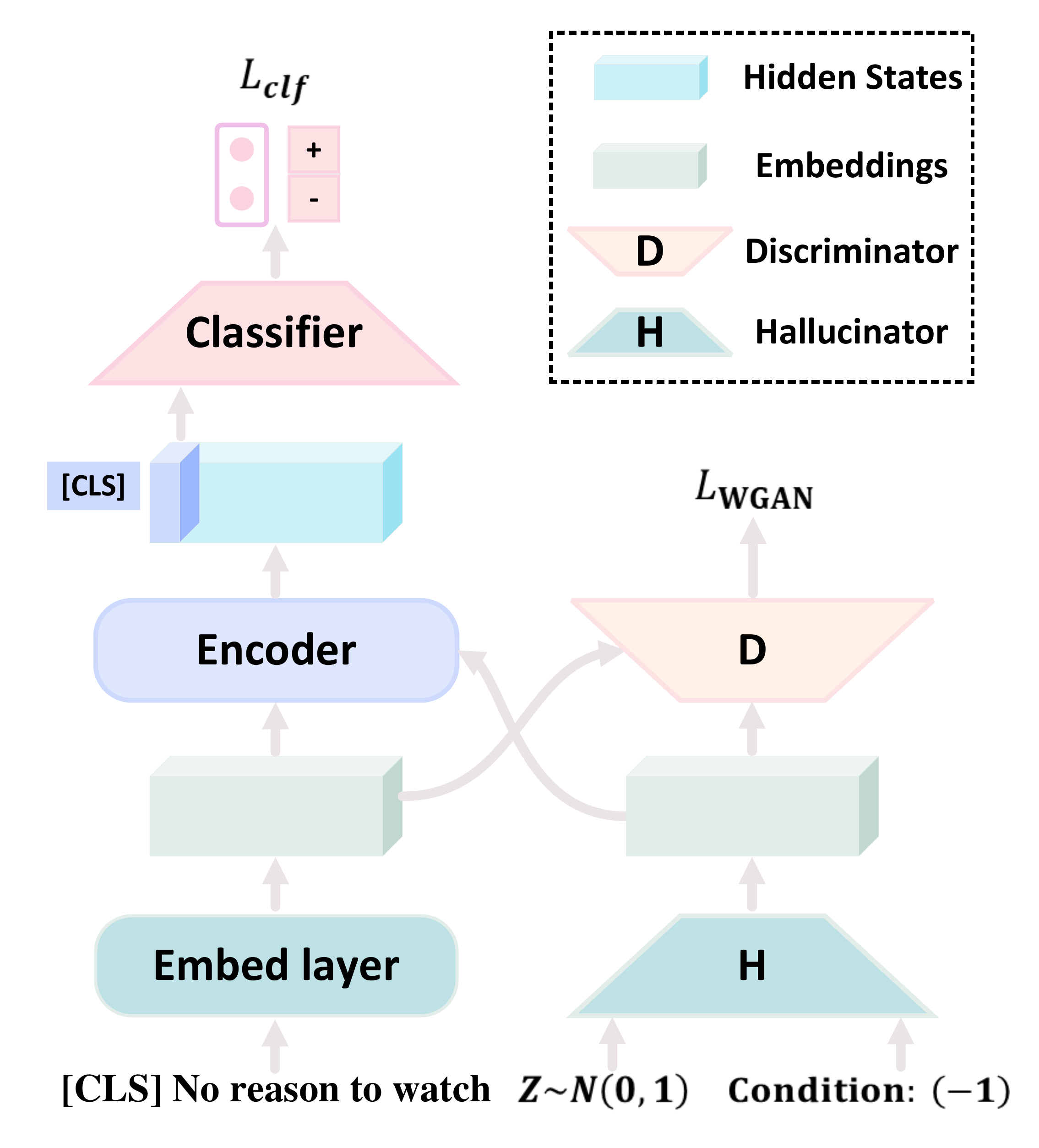}
\caption{Overview of our method: The encoder model takes real embeddings of sentences and hallucinated embeddings as input and learns from their mixture. Hallucinator $H$ generates fake embeddings conditioned on the class label. The discriminator $D$ discriminates real embeddings and fake embeddings leading to the GAN loss $\mathcal{L}_{WGAN}$.}
\label{fig:overview}
\vspace{-0.2in}
\end{figure}

\section{Method}
\subsection{Conditional Wasserstein GAN}
GAN \cite{GAN} has led the revolution of generative models to achieve impressive results in synthesizing images \cite{CycleGAN2017} and higher dimensional data \cite{wang2020enhanced}. Wasserstein GAN (WGAN) \cite{pmlr-v70-arjovsky17a} uses the Wasserstein distance as the objective function to stabilize the training of GAN.

Our hallucinator is trained under the conditional WGAN framework. After the training, we use it to generate pseudo-embeddings of examples by feeding it with random noisy vectors $z$ sampled from $\mathcal{N}(0,1)$ and the corresponding condition class labels $c_{i}$. The hallucinated embeddings $s_{\text{halluc}}$, in principal, are indiscriminative to the embeddings of observed examples in that class.

\subsection{Fine-tuning with Hallucinated Embedding}
For a single input sentence, we first pass it through the embedding layer to get the sentence embedding $s_{\text{sent}}$. We then concatenate $s_{\text{sent}}$ with $s_{\text{halluc}}(c_{i})$ to form a batch of mixture of real and fake embeddings $[s_{\text{sent}}, s_{\text{halluc}}(c_{i})]$. The encoder learns from the batch with the corresponding labels $[c_{\text{sent}}, c_{i}]$.

\textbf{Label Calibration}. The hallucinated embedding $s_{\text{halluc}}(c_{i})$ is conditioned on its label $c_{i}$. However, this hard label may not best represent the class information of the hallucinated embedding. We propose Label Calibration (LabelCalib) by pseudo-labeling from a teacher model $\mathcal{F}_{\text{GEN0}}$ ($LM_{1}$ in Algorithm \ref{alg:main}), where $\mathcal{F}_{\text{GEN0}}$ is first fine-tuned on the original training set (without augmentation). The soft-label of the embedding $s_{\text{halluc}}(c_{i})$ is then $c_{\text{pseudo},i}=\mathcal{F}_{\text{GEN0}}(s_{\text{halluc}}(c_{i}))$. Finally, the language model $\mathcal{M}$ learns from the hallucinated embedding by KL-divergence 
\begin{align}
    \mathcal{L}_\text{halluc}=\text{KL}(\mathcal{M}(s_{\text{halluc}}(c_{i})), c_{\text{pseudo},i})
\end{align}
The total loss of our method is 
\begin{align}
    \mathcal{L}_\text{total}=\mathcal{L}_\text{real} + \mathcal{L}_\text{halluc}
\end{align}
where $\mathcal{L}_\text{real}$ is the loss learning from real embedding-label pairs. The pseudo-code for fine-tuning of few-shot language learners with hallucinated embeddings is shown in Algorithm \ref{alg:main}.

Note that baselines considered in this paper use total loss $\mathcal{L}_\text{total}=\mathcal{L}_\text{real}$. Computing $\mathcal{L}_\text{halluc}$ requires one additional forward pass of the hallucinator and one more forward pass and backward pass of the language model. Thus, our method has about $\times2$ computational overhead compared to the baselines.

\begin{figure}[ht]
\centering
\begin{minipage}{1.0\linewidth}
\begin{algorithm}[H]
\caption{Our method: EmbedHalluc}\label{alg:main}
\begin{algorithmic}[1]
\State $Max\_Step = 1000$, \\
$LM$: Language model, \\
$H$: Emebedding hallucinator (pre-trained), \\
$Train\_Set$: Training set, \\ 
$Sample$: Randomly sampling function, \\
$CE$:  Cross Entropy loss, \\
$KL$:  KL-divergence loss.
\For{i in $Max\_Step$} {$\triangleright$ Training $\text{LM}_{1}$}
    \State $sent, y = Sample(Train\_Set)$
    \State {$output_{1} = LM_{1}(sent)$}
    \State {$L = CE(output_{1}, y)$}
    \State {$L.backward()$}
    \State {$optimizer.step()$}
\EndFor

\For{i in $Max\_Step$} {$\triangleright$ Training $\text{LM}_{2}$}

    \State $sent, y = Sample(Train\_Set)$
    \State $embed = H(\mathcal{N}(0,1)), c)$
    
    {$\triangleright$ Learning from real text}
    \State {$output_{1} = LM_{2}(sent)$}
    \State {$L_{real}= CE(output_{1}, y)$}
    \State {$L_{real}.backward()$}
    \State {$optimizer.step()$}
    
    {$\triangleright$ Learning from hallucination}
    \State  {$prob_{2} = LM_{1}(embed)$}
    \State  {$output_{2} = LM_{2}(embed)$}
    \State  {$L_{halluc} = KL(prob_{2}, output_{2})$}
    \State  {$L_{halluc}.backward()$}
    \State  {$optimizer.step()$}
\EndFor
\State \Return $LM_{2}$
\end{algorithmic}
\end{algorithm}

\end{minipage}
\end{figure}

\section{Experiments}
\subsection{Evaluation Datasets and Protocol}
We evaluate our method on 15 classification tasks. The evaluations are conducted by averaging results on 5 different train test splits. We sample 16 examples per class to form a training set and construct a validation set with the same size as the training set.

\subsection{Training Details for Embedding Hallucinators}
The training of Embedding Hallucinators involves training a generator and discriminator in the cWGAN framework. The generator is a 4-blocks model, with each block containing a FullyConnect layer followed by a BatchNorm and LeakyReLU. The hidden dimensions of the generator are $128,256,512,1024$. The hallucinated embeddings, i.e., outputs of the generator are tensors of $L \times 1024$, where the length of the generated embeddings $L$ is set to be 128. The discriminator is a 3-blocks model, each bock having a sequence of FullyConnect-BatchNorm-LeakyReLU with the same hidden dimension of $512$.

We train the Embedding Hallucinators for 150 epochs using a batch size of 64, the Adam optimizer ($\beta=(0.5,0.999)$), and a learning rate of 0.0002. The real embeddings are collected from the language few-shot training set by passing text into the embedding layer of the language model. We apply gradient penalty with weight of loss 100 for training the cWGAN.

\subsection{Training Details for Few-Shot Language Learners}
We draw two mini-batches during the training of our few-shot language learners, i.e., one from the real language few-shot training set, another one by sampling the hallucinators (see Algorithm \ref{alg:main}).

To fairly compare our method with baselines and other methods, when learning with real sentences, we use the same learning rate of $1e^{-5}$ (further justification of using this learning rate can be found in Appendix~\ref{section:basline-LR}). Our method learns from hallucinated embeddings with a grid search of learning rate of $1e^{-5}, 5e^{-6}, 1e^{-6}$, and batch size of $4, 6, 8$. We use the same search for EDA \cite{wei-zou-2019-eda} and semi-supervised pseduo-labeling (SSL) when learning with additional augmented or pseudo-labeled data.

The models are selected based on the validation accuracy every 100 steps. Finally, results are reported by testing the models on the testing dataset. The algorithm is implemented in PyTorch-1.10 and experiments are conducted on Nvidia RTX-6000 and RTX-A6000 GPU.

\subsection{Main Results on 15 Tasks}
We compare our method EmbedHalluc (w/o or w/ LabelCalib) using RoBERTa-large on 15 tasks with two fine-tuning methods: conventional (Table~\ref{table:roberta-finetune}) and prompt-based fine-tuning (Table~\ref{table:roberta-prompt}). Results for BERT-large-cased can be found in Appendix~\ref{sec:bert-embedhalluc}.

\begin{table}[ht]
\begin{center}
\scriptsize
\begin{tabular}{l c c c}
           \hline
           Task            &     Fine-tuning      &    EmbedHalluc    &     w/LabelCalib\\
           \hline
           SST-2 (acc) & 76.8 (4.2) & \textbf{82.6} (5.6) & 82.0 (4.7) \\
           Subj (acc) & 90.3 (1.5) & \textbf{91.3} (0.8) & \textbf{91.3} (0.9)\\
           SST-5 (acc) & 40.6 (2.2) & 40.3 (1.5) &  \textbf{41.6} (2.6) \\
           CoLA (Matt.) & 36.0 (9.9) & \textbf{39.7} (10.8) & 38.1 (11.8) \\
           TREC (acc) & 83.0 (4.9) & \textbf{88.1} (2.5)  & 87.9 (1.0) \\
           MNLI (acc) & 41.6 (5.2) & 48.0 (9.5)  &  \textbf{49.6} (5.8) \\
           MNLI-mm (acc) & 42.7 (5.9) &  49.7 (10.5) & \textbf{51.8} (6.1)\\
           SNLI (acc) & 52.9 (6.7) & \textbf{54.4} (3.4) &  52.3 (5.3) \\
           QNLI (acc) & 55.3 (2.7) & 60.2 (5.3) &  \textbf{64.9} (5.1) \\
           QQP (acc) & 59.2 (8.6) &  64.6 (5.0) &  \textbf{66.7} (5.3) \\
           RTE (acc) & 52.9 (1.4) & 53.4 (1.7) & \textbf{55.9} (4.3)   \\
           MRPC (F1)  & 76.3 (5.2) & \textbf{78.7} (1.9)  & 78.1 (3.0) \\
           MR (acc)  & 74.5 (5.9) & 79.4 (5.5)  & \textbf{80.8} (3.2) \\
           MPQA (acc) & 65.0 (1.5) & 70.1 (7.0) &  \textbf{70.5} (4.6) \\
           CR (acc)  & 71.7 (7.5) & 75.1 (5.6) &  \textbf{78.0} (3.8)   
           \\ \hline

\end{tabular}
\caption{Comparison of conventional fine-tuning and our EmbedHalluc, using RoBERTa-large. Our Label Calibration (LabelCalib) can further improve the results.}
\label{table:roberta-finetune}
\end{center}
\end{table}

\begin{table}[ht]
\begin{center}
\scriptsize

\begin{tabular}{l c c c}
           \hline
           Task            &     Prompt-based      &    EmbedHalluc    &     w/LabelCalib \\
           \hline
           SST-2 (acc).  &  92.7 (0.4) &  92.8 (0.7)  &    \textbf{93.1} (0.7) \\
           Subj (acc)    & 91.3 (1.0)  &  \textbf{92.0} (0.4) &   91.7 (1.3)  \\
           SST-5 (acc)   & 48.8 (1.0)  & 49.0 (2.2) &     \textbf{49.4} (1.4)\\
           CoLA (Matt.)  &  7.3 (5.8) &  12.3 (7.6) &    \textbf{22.1} (15.6) \\
           TREC (acc)    & 83.8 (5.3)  &  85.5 (3.3) &    \textbf{87.1} (2.9)\\
           MNLI (acc)    & \textbf{69.7} (2.0)  & 68.0 (2.8)  &     68.5 (1.7)\\
           MNLI-mm (acc) & \textbf{71.5} (1.9)  & 69.9 (3.0) & 70.6 (1.7)   \\
           SNLI (acc)    & 78.0 (3.0)  &  \textbf{78.8} (2.3) &    78.4 (2.3)\\
           QNLI (acc)    & 68.6 (2.8)  & 69.6 (0.3)  &    \textbf{71.6} (2.0) \\
           QQP (acc)     & 70.2 (4.3)  &  71.9 (5.2) & \textbf{74.2} (0.9)\\
           RTE (acc)     & \textbf{70.9} (3.3)  &  69.9 (3.3)  &  66.9 (3.4)   \\
           MRPC (F1)     & 74.6 (6.8)  &   78.0 (4.9)  &    \textbf{80.3} (3.5)  \\
           MR (acc)      & 86.8 (0.9)  &   87.2 (0.9)  &   \textbf{87.5} (0.9) \\
           MPQA (acc)    & 85.4 (1.8)  & 84.2 (1.9)  &    \textbf{85.4} (1.9)\\
           CR (acc)      & 91.1 (1.0)  & 91.1 (0.9)   &   \textbf{91.3} (0.3)  
           \\ \hline

\end{tabular}
\caption{Comparison of prompt-based fine-tuning and our EmbedHalluc, using RoBERTa-large.}
\label{table:roberta-prompt}
\end{center}
\end{table}

In conventional fine-tuning, EmbedHalluc improves over the baseline in 14 tasks, only marginally under-performs in SST-5 (40.3 vs. 40.6 of baseline). When combining with LabelCalib, our method outperforms in all tasks. When applying to prompt-based fine-tuning, while our method under-performs in MNLI, MNLI-mm and RTE, it outperforms for all other tasks, with substantial improvements over the baseline in CoLA, TREC, QNLI, MRPC.

The relatively smaller improvements for prompt-based methods
may be due to the  inconsistency and randomness in the learning process since we have to insert $\texttt{[mask]}$ token to a random position in the hallucinated embedding $s_\text{halluc}$, for the calculation of the loss. Whereas, in conventional fine-tuning, the $\texttt{[CLS]}$ token is always appended to the beginning of $s_\text{halluc}$ and the classification is performed at the $\texttt{[CLS]}$ token. 

\subsection{Comparing to EDA and SSL}
Since our method is a generative data augmentation (DA) method, we compare it to another DA method EDA. We also consider semi-supervised learning (SSL) which relies on unlabeled data (64 examples per class in our experiments). We apply pseudo-labeling \cite{cascante2021curriculum} for SSL, \ie, we first fine-tune the model with the few-shot training set and use the fine-tuned model to pseudo-label the unlabeled data, finally we fine-tune the model again with the few-shot training set combined with the pseudo-labeled set.

EDA edits the input sentences by applying synonym replacement, random swap, random deletion and random insertion for a default 10\% ($\alpha$) of tokens. EDA either greatly change the sentence with a large $\alpha$ or fails to introduce substantial variations (which is crucial in the extreme low data setting) of inputs with a small $\alpha$. Since it operates in the continuous embedding space, EmbedHalluc hallucinates diverse embeddings that follow the distribution of few-shot set. Thus, we observe in Table~\ref{table:roberta-embedhalluc-eda-semi} that EmbedHalluc is overall superior to EDA. 

EmbedHalluc is still competitive when comparing against SSL which assumes to have additional 64 examples per class from the task distribution.

\begin{table}[ht]
\begin{center}
\scriptsize

\begin{tabular}{l c c c c}
           \hline
           Task     &      fine-tuning       &     EmbedHalluc      &    EDA    & SSL \\
           \hline
           SST-2  & 76.8 (4.2)  & 82.6 (5.6) & 82.3 (6.2)  & \textbf{83.2} (6.0) \\
           Subj & 90.3 (1.5) & \textbf{91.3} (0.8) & 89.2 (2.0) &  91.2 (1.0)\\
           SST-5  &  40.6 (2.2) & 40.3 (1.5) &   38.8 (3.7)   &  \textbf{41.7} (1.9)\\
           CoLA  & 36.0 (9.9)  & \textbf{39.7} (10.8) &  25.5 (11.0)  &  39.6 (11.8)\\
           TREC   &  83.0 (4.9) & \textbf{88.1} (2.5)  &  84.0 (1.9)   &  87.4 (3.4)\\
           MNLI & 41.6 (5.2)  & \textbf{48.0} (9.5)  &  42.0 (3.9)  &  43.9 (4.2)\\
           MNLI-mm  &  42.7 (5.9) &  \textbf{49.7} (10.5) & 44.2 (3.4)  &  45.6 (4.7)\\
           SNLI  &  52.9 (6.7) & 54.4 (3.4) &  48.0 (4.7) &  \textbf{54.9} (7.7)\\
           QNLI  & 55.3 (2.7)  & \textbf{60.2} (5.3) &   58.7 (5.3)   &  53.6 (1.8)\\
           QQP & 59.2 (8.6)  &  \textbf{64.6} (5.0) &   60.7 (6.8)   &  63.2 (7.1)\\
           RTE & 52.9 (1.4)  & 53.4 (1.7) &   53.0 (4.9)    &  \textbf{53.9} (1.2)\\
           MRPC  & 76.3 (5.2)  & \textbf{78.7} (1.9)  &  73.8 (7.5)  &  77.3 (5.4)\\
           MR  & 74.5 (5.9)  & \textbf{79.4} (5.5)  &  78.1 (2.5)   &  77.9 (4.9)\\
           MPQA & 65.0 (1.5)  & 70.1 (7.0) &  \textbf{72.8} (7.8)   & 68.8 (3.5) \\
           CR   & 71.7 (7.5)  & 75.1 (5.6) &   \textbf{80.7} (5.2)   &  75.6 (8.6)
           \\ \hline

\end{tabular}
\caption{Comparison of EmbedHalluc, EDA, and SSL by pseudo-labeling, using RoBERTa-large as the base model and conventional fine-tuning as the base learning method.}
\label{table:roberta-embedhalluc-eda-semi}
\end{center}
\end{table}

\subsection{Negative Results from Regularizations}
Our method can also be viewed as an implicit regularization method. Thus, we also compare to two latest methods for better fine-tuning language models with regularization. \citet{zhang2021revisiting} find that fine-tuning can be achieved by: correcting bias in the optimizer, re-initialization of top layers, and training longer. Correcting bias in the optimizer is already fixed by the default optimizer in Huggingface Transformer and training longer surely will lead to further over-fitting in our extreme data scarce scenario. Thus, we consider re-initialization (Re-Init) of top layers as one of our comparisons. We further compare against Mixout \cite{lee2019mixout}, which is shown to be an effective regularization when fine-tuning with a few thousand examples. We used the public code for both of these methods. Since we adapt their code to our extreme data deficient setting, we re-search the hyper-parameters of both methods (including their suggested values). For Re-Init, we search the top 1,2,3,4,5 layers; and for Mixout, we search mixout rate from $0.1,0.2,...,0.9$ and report their best results in Table~\ref{table:roberta-embedhalluc-reinit-mixout}, using RoBERTa-large. Results for BERT-large-cased can be found in Appendix~\ref{sec:roberta-reinit-mixout}.

We find that those two methods fail to alleviate the over-fitting problem in such extreme setting, though they have been to be effective when given a few thousands examples.

\begin{table}[!ht]
\begin{center}
\scriptsize
\begin{tabular}{l c c c c}
           \hline
           Task & fine-tuning  & EmbedHalluc &  Re-init  &  Mixout  \\
           \hline
           SST-2  & 76.8 (4.2) &  \textbf{82.6} (5.6) &  82.5 (1.9)  & 78.5 (9.4) \\
           Subj & 90.3 (1.5) &  \textbf{91.3} (0.8) &  91.1 (2.4)  & 90.3 (0.8) \\
           SST-5  &  40.6 (2.2) & 40.3 (1.5) &   \textbf{41.2} (1.9)  & 37.5 (3.0)\\
           CoLA  & 36.0 (9.9)  &  \textbf{39.7} (10.8) &  33.4 (8.1)  & 38.6 (5.9) \\
           TREC   &  83.0 (4.9) &  \textbf{88.1} (2.5)  &  81.8 (5.6) & 86.0 (3.4)  \\
           MNLI & 41.6 (5.2)  &  \textbf{48.0} (9.5)  & 43.7 (4.3)   &  42.7 (4.6) \\
           -mm  &  42.7 (5.9) &   \textbf{49.7} (10.5) & 45.2 (4.8)   & 45.0 (5.4) \\
           SNLI  &  52.9 (6.7) &  \textbf{54.4} (3.4) &  52.1 (2.2)  &  53.7 (3.8)  \\
           QNLI  & 55.3 (2.7)  &  \textbf{60.2} (5.3) &  59.8 (5.0)  & 57.1 (3.5)  \\
           QQP & 59.2 (8.6)  &  \textbf{64.6} (5.0) &  60.2 (10.6)  &  62.4 (6.0)  \\
           RTE & 52.9 (1.4)  & 53.4 (1.7) &   52.5 (5.4) &   \textbf{53.5} (2.5)  \\
           MRPC  & 76.3 (5.2)  &  \textbf{78.7} (1.9)  &  67.0 (20.1)  & 77.4 (2.7)\\
           MR  & 74.5 (5.9)  &  \textbf{79.4} (5.5)  & 71.3 (9.0)   & 67.6 (10.0) \\
           MPQA & 65.0 (1.5)  &  \textbf{70.1} (7.0) &  68.8 (7.5)  & 68.0 (5.6)  \\
           CR   & 71.7 (7.5)  & 75.1 (5.6) &   \textbf{83.0} (2.2)  &  67.5 (5.4)    
           \\ \hline

\end{tabular}
\caption{Comparisons of EmbedHalluc to Re-init and Mixout, using RoBERTa-large as base models and conventional fine-tuning as the base learning method.}
\label{table:roberta-embedhalluc-reinit-mixout}
\end{center}
\end{table}

\section{Comparing to Adversarial Training}
Adversarial training adds noise into the training data to increase the robustness of a model. It has been shown that adversarial training can also improve the performance of language models. Here, we compare EmbedHalluc to two recent adversarial training methods, freeLB \cite{Zhu2020FreeLB:} and SMART \cite{jiang-etal-2020-smart} adapted to our setting. For freeLB, we use the publicly available code and suggested hyper-parameters for each task. In addition to the default batch size and learning rate used in the baseline fine-tuning and EmbedHalluc, we also search additional batch sizes and learning rates for freeLB. We use the default setting for SMART. As shown in Table~\ref{table:roberta-embedhalluc-freeLB-SMART}, with one exception, our method largely outperforms freeLB and SMART.

\begin{table}[!ht]
\begin{center}
\small
\begin{tabular}{l c c c}
           \hline
           Task  & EmbedHalluc &  freeLB  & SMART \\
           \hline
           SST-2  &  82.6 (5.6)  & 78.5 (8.8)  & \textbf{83.6} (3.1) \\
           CoLA   &  \textbf{39.7} (10.8) & 31.6 (11.1) & 34.2 (4.1) \\
           MNLI   &  \textbf{48.0} (9.5)  & 40.8 (3.5)  & 39.1 (4.1) \\
           -mm    &  \textbf{49.7} (10.5) & 42.1 (4.3)  & 40.0 (4.9)\\
           QNLI   &  \textbf{60.2} (5.3)  & 58.2 (5.0)  & 55.5 (2.9)\\
           QQP    &  \textbf{64.6} (5.0)  & 62.4 (3.8)  & 56.5 (6.1)\\
           RTE    &  \textbf{53.4} (1.7)  & 52.3 (4.2)  & 49.2 (2.6)\\
           MRPC   &  \textbf{78.7} (1.9)  & 76.8 (3.5)  & 77.1 (3.1)
           \\ \hline

\end{tabular}
\caption{Comparisons of EmbedHalluc to freeLB and SMART, using RoBERTa-large as base models and conventional fine-tuning as the base learning method.}
\label{table:roberta-embedhalluc-freeLB-SMART}
\end{center}
\end{table}

\section{Limitations}
While EmbedHalluc works well empirically, it relies on hallucinating non-interpretable embeddings to facilitate the learning process. Besides, the learning of cWGAN requires careful human attention to maintain a stable training.

\section{Conclusion}
In this paper, we introduce an embedding hallucination method for data augmentation for few-shot learning, based on cWGAN. The proposed method improves over the baselines in 15 tasks and outperforms a common augmentation method, and two recent regularization methods.

\section{Ethics Statement}
As far as we are aware, our proposed work does not have any explicit ethical concerns. However, our work relies on pre-trained language models, which have been shown to be biased in prior work \cite{liang2021towards}. As such, users of such models, specially for sensitive applications, should be aware of and if possible address such issues.

\bibliography{custom}
\bibliographystyle{acl_natbib}

\clearpage
\appendix

\section{Best Learning Rate for RoBERTa-prompt}
\label{sec:LR}
Here, we provide best learning rates (LR, searched from $1e^{-5}, 5e^{-6}, 1e^{-6}$ as discussed in main paper) for $\mathcal{L}_\text{halluc}$ of EmbedHalluc for each task used in RoBERTa-large prompt-based fine-tuning.

\setcounter{table}{0}
\renewcommand\thetable{\Alph{section}.\arabic{table}}

\begin{table}[ht]
\begin{center}
\scriptsize

\begin{tabular}{l  c}
           \hline
           Task               &  LR   \\
           \hline
           SST-2     &  $1e^{-6}$ \\
           Subj      &  $1e^{-5}$ \\
           SST-5     &  $1e^{-6}$ \\
           CoLA      &  $1e^{-5}$ \\
           TREC      &  $1e^{-6}$ \\
           MNLI      &  $1e^{-5}$ \\
           MNLI-mm   &  $1e^{-5}$ \\
           SNLI      &  $1e^{-6}$ \\
           QNLI      &  $5e^{-6}$ \\
           QQP       &  $1e^{-6}$ \\
           RTE       &  $1e^{-6}$ \\
           MRPC      &  $1e^{-6}$ \\
           MR        &  $5e^{-6}$ \\
           MPQA      &  $5e^{-6}$ \\
           CR        &  $5e^{-6}$ 
           \\ \hline

\end{tabular}
\caption{Best learning rate for $\mathcal{L}_\text{halluc}$ found for RoBERTa-large prompt-based fine-tuning.}
\label{table:HyperParams}
\end{center}
\end{table}

\section{EmbedHalluc with BERT}\label{sec:bert-embedhalluc}
In addition to the experiments using RoBERTa shown in the main paper, here we show the results of BERT-large-cased with conventional fine-tuning as a further check on robustness of our method with respect to the choice of model. Table~\ref{table:bert-finetune} shows the results of the experiments. EmbedHalluc outperforms the baseline across 14 of the 15 tasks with an average improvement of 2.43 over the baseline.

\setcounter{table}{0}
\renewcommand\thetable{\Alph{section}.\arabic{table}}
\begin{table}[ht]
\begin{center}
\scriptsize
\begin{tabular}{l c c }
           \hline
           Task            &     fine-tuning     &    EmbedHalluc    \\
           \hline
           SST-2 (acc)  & 73.9 (5.4)   &  \textbf{76.6} (3.8) \\
           Subj (acc)   & 85.2 (3.4) & \textbf{89.0} (0.9)  \\
           SST-5 (acc) &  37.6 (4.5) & \textbf{38.9} (4.2) \\
           CoLA (Matt.) &  21.9 (10.0) & \textbf{28.7} (7.2) \\
           TREC (acc)   & 77.6 (6.3) &  \textbf{81.8} (3.1)  \\
           MNLI (acc)   &   35.5 (0.9) &  \textbf{36.1} (0.8) \\
           MNLI-mm (acc)  & 36.0 (0.9)  & \textbf{36.7} (1.5) \\
           SNLI (acc)   & 39.7 (3.6)  & \textbf{41.0} (3.2) \\
           QNLI (acc)    & 53.9 (2.5)  & \textbf{55.1} (3.0) \\
           QQP (acc)    & 56.7 (4.3) & \textbf{59.5} (2.9) \\
           RTE (acc)    & 50.9 (3.5)  & \textbf{54.2} (2.5) \\
           MRPC (F1)    & 72.3 (6.5)& \textbf{76.7} (2.8) \\
           MR (acc)     &   \textbf{73.2} (6.3) &  66.4 (8.9) \\
           MPQA (acc)   & 64.6 (5.6)  &  \textbf{65.3} (4.0) \\
           CR (acc)     &  67.9 (9.0) &   \textbf{77.5} (12.6)  
           \\ \hline

\end{tabular}
\caption{Comparison of conventional fine-tuning and our EmbedHalluc, using BERT-large-cased.}
\label{table:bert-finetune}
\end{center}
\end{table}

\section{Regularization Methods with BERT}
\label{sec:roberta-reinit-mixout}
Besides the experiments with RoBERTa-large shown in the main paper, we present Re-Init and Mixout using BERT-large-cased in this section. The results are shown in Table \ref{table:bert-embedhalluc-reinit-mixout}. 

Qualitatively similar to what we observe with experiments using RoBERTa-large in the main paper, Re-Init and Mixout fail to outperform EmbedHalluc in most tasks, with the exceptions of SNLI and QNLI.

\setcounter{table}{0}
\renewcommand\thetable{\Alph{section}.\arabic{table}}
\begin{table}[!ht]
\begin{center}
\scriptsize

\begin{tabular}{l c c c c}
           \hline
           Task      &      fine-tuning      &     EmbedHalluc      &    Re-init    &     Mixout \\
           \hline
          SST-2   & 73.9 (5.4)  &\textbf{76.6} (3.8) & 73.6 (4.2) &  71.9 (3.9) \\
           Subj   & 85.2 (3.4)  &\textbf{89.0} (0.9)  & 87.0 (2.5) & 85.6 (1.1) \\
           SST-5  & 37.6 (4.5)  &\textbf{38.9} (4.2) & 36.4 (2.5) & 36.0 (3.1) \\
           CoLA   & 21.9 (10.0)  &\textbf{28.7} (7.2) & 23.8 (11.8)  & 13.0 (11.2) \\
           TREC    &  77.6 (6.3)  &\textbf{81.8} (3.1)  & 79.7 (3.4) & 78.8 (4.4) \\
           MNLI     &  35.5 (0.9)  &\textbf{36.1} (0.8) & 35.1 (1.6) & 33.2 (0.6)\\
           MNLI-mm    & 36.0 (0.9)  &\textbf{36.7} (1.5) & 35.6 (2.2) & 33.6 (1.0) \\
           SNLI    &  39.7 (3.6) &41.0 (3.2) & \textbf{45.3} (3.3) & 42.5 (3.2) \\
           QNLI      & 53.9 (2.5)  &55.1 (3.0) & \textbf{55.9} (2.0) & 55.0 (2.7) \\
           QQP      &  56.7 (4.3) &\textbf{59.5} (2.9) & 58.9 (2.9) & 56.6 (6.1)\\
           RTE      & 50.9 (3.5)  &\textbf{54.2} (2.5) & 51.5 (3.8) & 50.8 (1.8) \\
           MRPC     & 72.3 (6.5)  &\textbf{76.7} (2.8) & 63.4 (4.6) & 74.1 (3.2) \\
           MR       &   \textbf{73.2} (6.3)  & 66.4 (8.9) & 60.8 (4.4) & 63.0 (4.5) \\
           MPQA    &  64.6 (5.6)  &\textbf{65.3} (4.0) &  64.8 (5.8)& 60.9 (2.9) \\
           CR      &  67.9 (9.0)   &\textbf{77.5} (12.6) & 71.6 (8.3) &  75.9 (6.1)
           \\ \hline

\end{tabular}
\caption{Comparisons of EmbedHalluc to Re-init and Mixout, using BERT-large as the base models and conventional fine-tuning as the base learning method.}
\label{table:bert-embedhalluc-reinit-mixout}
\end{center}
\end{table}

\section{Learning Rate for Baselines} \label{section:basline-LR}
The baseline has only one loss $\mathcal{L}_{\text{real}}$, whereas we are learning with an additional loss $\mathcal{L}_{\text{halluc}}$, making the total loss to be $\mathcal{L}_{\text{real}} + \mathcal{L}_{\text{halluc}}$. The learning rate for $\mathcal{L}_{\text{real}}$ in the baselines and ours are kept the same. Note that we do not search for this learning rate for our method. We choose $1e^{-5}$, which is the most common learning rate to finetune BERT/RoBERTa. As we show in Table~\ref{table:LMBFF-lr}, this learning rate produces reasonably good results for the baselines, being the best for 13 tasks and only marginally under-performing in the other 2 tasks. The results in Table~\ref{table:LMBFF-lr} are generated by running the baselines with a batch size of 2 and different learning rates $1e^{-5}$, $2e^{-5}$, $5e^{-5}$ suggested by \citet{gao2021making}. 

\setcounter{table}{0}
\renewcommand\thetable{\Alph{section}.\arabic{table}}
\begin{table}[!ht]
\begin{center}
\scriptsize
\begin{tabular}{lccc}
\hline
prompt & 1e-5   & 2e-5   & 5e-5     \\
\hline
SST-2               & \textbf{92.7} (0.4) &  91.0 (2.4) & 83.6 (7.8)   \\
subj                & \textbf{91.3} (3.3) & 87.8 (3.3) & 83.9 (3.0)   \\
SST-5               &  \textbf{48.8} (1.0) & 48.1 (1.0) & 43.6 (1.9)   \\
CoLA                & 7.3 (5.8)  & 8.7 (5.7)  &  \textbf{10.5} (7.2)    \\
trec                &  \textbf{83.8} (5.3) & 79.3 (5.2) & 78.0 (0.7)   \\
MNLI                &  \textbf{69.7} (2.0) & 66.5 (3.2) & 60.0 (5.2)   \\
MNLI-mm             &  \textbf{71.5} (1.9)  &  68.7 (3.2) & 62.9 (5.3)   \\
SNLI                &  \textbf{78.0} (3.0) & 76.2 (2.3)) & 54.7 (10.3)   \\
QNLI                &  \textbf{68.6} (2.8) & 64.9 (3.1) & 63.0 (7.2)   \\
QQP                 &  \textbf{70.2} (4.3) & 64.2 (5.0) & 57.7 (4.9)   \\
RTE                 &  \textbf{70.9} (3.3) & 63.3 (4.8) & 55.8 (8.4)   \\
MRPC                &  \textbf{74.9} (6.8) & 73.9 (3.8) & 72.8 (2.6)  \\
MR                  &  \textbf{86.8} (0.9) & 84.5 (1.8) & 80.1 (3.9)   \\
MPQA                & 85.4 (1.8) &  \textbf{85.6} (1.1) & 80.1 (3.9)   \\
CR                  &  \textbf{91.1} (1.0) & 90.6 (0.9) & 85.3 (2.6)  \\
\hline
\end{tabular}
\caption{Results of baseline with Roberta-large prompt-based fine-tuning on different learning rate. $1e^{-5}$ is what we used in main experiments.
}
\label{table:LMBFF-lr}
\end{center}
\end{table}

\end{document}